\documentclass[10pt,twocolumn,letterpaper]{article}

\usepackage{cvpr}
\usepackage{times}
\usepackage{epsfig}
\usepackage{graphicx}
\usepackage{amsmath}
\usepackage{amssymb}

\usepackage[pagebackref=true,breaklinks=true,letterpaper=true,colorlinks,
bookmarks=false]{hyperref}

\cvprfinalcopy 

\def\cvprPaperID{1924} 

\ifcvprfinal\fi
\begin{document}

\title{TemplateNet for Depth-Based Object Instance Recognition}

\author{Ujwal Bonde\\
University of Cambridge\\
{\tt\small ub216@cam.ac.uk}
\and
Vijay Badrinarayanan\\
University of Cambridge\\
{\tt\small vb292@cam.ac.uk}
\and
Roberto Cipolla\\
University of Cambridge\\
{\tt\small rc10001@cam.ac.uk}
\and
Minh-Tri Pham\\
AIG\\
{\tt\small minhtri.pham@aig.com}
}

\maketitle

\begin{abstract}
We present a novel deep architecture termed \textit{templateNet} for depth based
object instance recognition. Using an intermediate \textit{template layer} we
exploit prior knowledge of an object's shape to sparsify the
feature maps. This has three advantages: (i) the network is better regularised
resulting in structured filters; (ii) the sparse feature maps results in
intuitive features been learnt which can be visualized as the output of the
template layer and (iii) the resulting network achieves state-of-the-art
performance. The network benefits from this without any additional
parametrization from the template layer. We derive the weight updates needed to
efficiently train this network in an end-to-end manner. We benchmark the
templateNet for depth based object instance recognition using two publicly
available datasets. The datasets present multiple challenges of clutter, large
pose variations and similar looking distractors. Through our experiments we show
that with the addition of a template layer, a depth based CNN is able to
outperform existing state-of-the-art methods in the field.

\end{abstract}

\section{Introduction}
Recent advances in computer vision has led to deep learning algorithms matching
and outperforming humans in various recognition tasks like digit
recognition~\cite{srivastava2014dropout}, face
recognition~\cite{conf/cvpr/TaigmanYRW14} and category
recognition~\cite{journals/corr/IoffeS15}. A common theme among them is
the use of additional techniques to better regularise the training of a network
with large number of parameters. For instance, using dropout, Srivastava \etal
prevent co-adaptation and overfitting~\cite{srivastava2014dropout}. Similarly,
using batch normalization, Ioffe \& Szegedy~\cite{journals/corr/IoffeS15} reduce
the degrees of freedom in the weight space. This encourages us to explore more
ways to regularise such large models by designing problem-specific
regularizers.

A popular approach for regularisation is to introduce sparsity into the
model~\cite{olshausen1997sparse}. However existing schemes of introducing
sparsity in deep networks generally cannot be trained in a end-to-end manner and
require a complex alternating schemes of
optimization~\cite{poultney2006efficient,lee2008sparse}. Moreover, in most of
these methods (including dropout) the sparse feature maps lack spatial
structure. Based on the application domain it might be possible to leverage
prior knowledge to introduce structure to the sparsity. Depth-based instance
recognition is one such application where we have a complete 3D scan of an
object during training and the goal is to recognize it under different
viewpoints in challenging 2.5D test scenes. We take advantage of this prior
information by employing an intermediate \textit{template layer} to introduce
structure to the sparse feature maps. We call this new
architecture \textit{TemplateNet} and derive weight update equations to train it
in an end-to-end manner. We show that with the additional regularisation of the
template layer, templateNet outperforms existing state-of-the-art methods in
depth based instance recognition.

In addition to introducing sparsity, observing the output of a template layer
allows us to visualize the learnt features of an object. This is a more natural
way of visualizing learnt features as compared to existing methods which either
use redundant layers~\cite{zeiler2010deconvolutional} that are not useful for
inference or need to backpropagate errors while maximizing class probability
using additional constrains to obey natural image
statistics~\cite{simonyan2013deep}.

\noindent In summary our main contributions in this paper are:
\begin{itemize}
 \item For the task of instance recognition, using a template layer we
impose structure on the sparse activations of feature maps. This regularises
the network and improves its performance without additional
parametrization.
 \item The output of template layer can be used to visualize the learnt features
for an object.
\end{itemize}
\begin{figure*}[t!]
    \centering   
\includegraphics[width=1\linewidth]{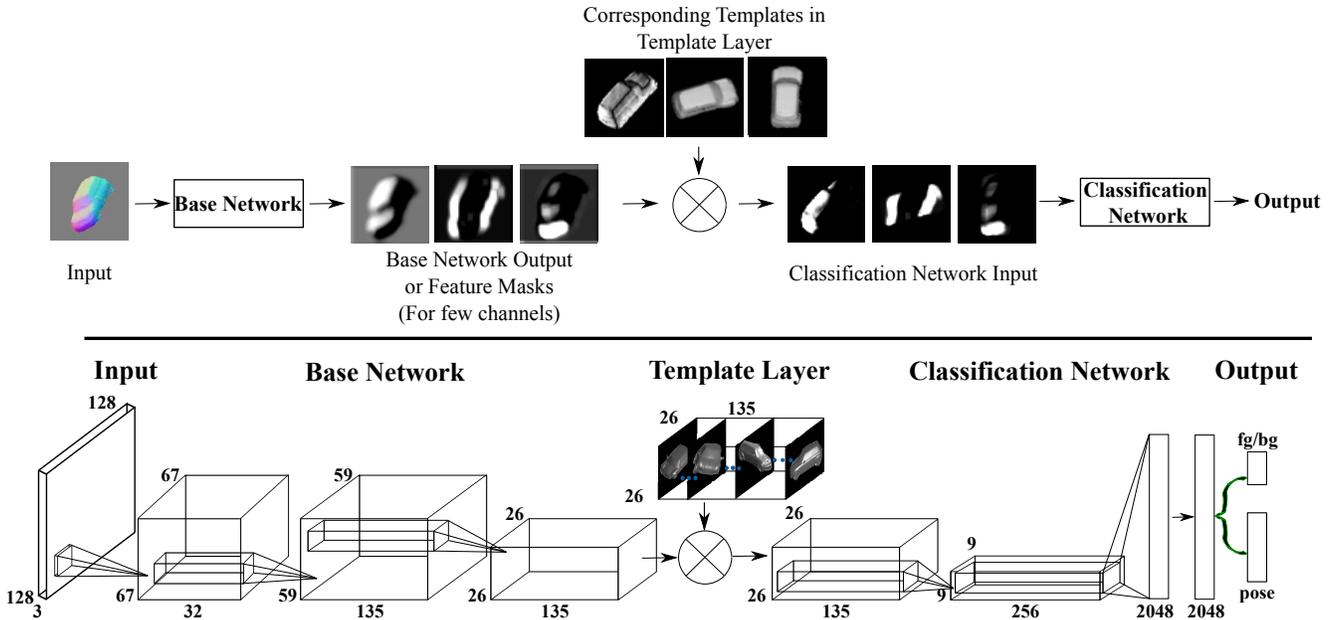}
    \caption{{\textbf{Top} pane: A block diagram view of our proposed
architecture. TemplateNet contains three modules (see
section~\ref{sec:templateNet}): (a) \textit{Base network}: extracts the feature
masks; (b) \textit{Template layer}: introduces sparsity to the feature masks
using prior knowledge; (c) \textit{Classification network}: outputs the final
predictions. TemplateNet allows us to visualize these intermediate feature masks
giving an intuition of the learnt features. For more visualization results
please see figure~\ref{fig:templateResponse}. \textbf{Bottom} pane: The final
architecture that we use in our experiments. The numbers quoted correspond to
the image/feature map dimensions and the number of channels used.}} 
    \label{fig:architecture}
\end{figure*}
\section{Literature Review}\label{sec:chpt3_litReview}
The advancement in sensor
technology~\cite{litomisky2012consumer,golrdon20123d} coupled with real time
reconstruction systems~\cite{conf/ismar/NewcombeIHMKDKSHF11} has resulted in
depth data being readily available. Encouraged by this several methods on
instance
recognition using depth data have been proposed. One of the most popular is that
of Hinterstoisser \etal~\cite{hinterstoisser2011multimodal}
which uses surface normals from depth images together with edge orientations
from RGB images as template features. Using a fast matching scheme the
authors match thousands of such template features from different viewpoints of
the object to robustly detect the presence of an object. However the lack of
discriminative training leads to poor performance in presence of similar looking
clutter.

Brachmann et al.~\cite{brachmann2014learning}, use a random forest to perform a
per pixel object pose prediction. Using an energy minimization scheme they
compute the final pose and location of an object in the scene. In contrast to
their multi-stage approach we use end-to-end learning to detect both the
location and pose of an object. In~\cite{tejani2014latent} rather than using
discriminative training, Tejani \etal use co-training with hough forest. This
avoids the need for background/negative training data. However during testing it
requires multiple passes over the trained forest to predict the location and
pose of the object. Moreover, in the absence of negative data it is unclear how
this method can perform in the presence of similar looking clutter.
In~\cite{conf/eccv/BondeBC14} using soft labelled random forest
(slRF)~\cite{criminisi2013decision}, Bonde
\etal perform discriminative learning on the manually designed features
of~\cite{hinterstoisser2011multimodal} and show impressive performance under
heavy as well as similar looking clutter using only the depth data. With
CNN's~\cite{Lecun1998} driving recent advances in computer vision we explore its
use in depth based instance recognition for performing feature learning. As our
experiments show (section~\ref{sec:chpt3_exp}), we need to regularise the CNN to
compete with
existing methods.

Sparsity has been widely used for better regularisation in both shallow
architectures~\cite{olshausen1997sparse} as well as in deep
architectures~\cite{lee2008sparse}. A popular approach to enforce sparsity in
deep architectures is to use an $L1$ norm penalty on the filter weights. Ranzato
\etal showed that better structured filters are obtained by
enforcing sparsity on the output of the filters (or feature maps) rather than
the filter weights~\cite{poultney2006efficient}. To achieve this they use a
\textit{sparsifying logistic} which converts the intermediate feature maps to a
sparse vector. However this results in a non-trivial cost function requiring a
complex alternating optimization scheme for computing weight updates. 
We propose an alternate manner of introducing sparsity by using the template
layer. We exploit the nature of the instance recognition problems by using prior
knowledge of the object shape to introduce structure to the sparse activations
of feature maps. This results in weight updates that can be easily computed for
the entire network using the chain rule, thus allowing us to train the network
in an end-to-end manner (section~\ref{sec:learn}). Our proposed network
outperforms existing methods on challenging publicly available datasets.

\section{TemplateNet}\label{sec:templateNet}
Figure~\ref{fig:architecture} (top pane) presents the architecture of
templateNet. The templateNet essentially contains three components: the base
network, the template layer and the classification network. The base network and
the classification network contain standard convolutional layers (or fully
connected layers). On the other hand, a template layer is essentially an element
wise multiplicative layer having one-to-one connection with the base networks
output. To better motivate the function of each component in a templateNet we
first draw connections to existing methods in literature for instance
recognition.

\begin{figure}[t]
    \centering
    \includegraphics[width=1\linewidth]{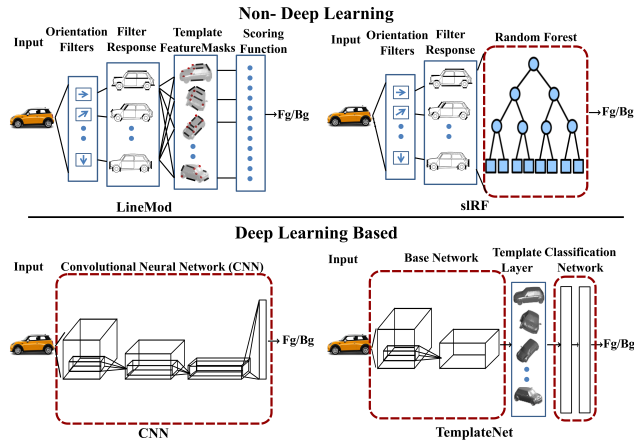}
    \caption{ {Block diagram view of different approaches to instance
recognition. \textbf{Top left} is the block diagram for Template Matching
(LineMod~\cite{hinterstoisser2011multimodal}) which uses manually designed
filter banks and distinguishing feature (or feature masks) giving a better
understanding of the system. \textbf{Top right} is the
discriminative method using slRF~\cite{conf/eccv/BondeBC14}. The discriminative
features are learnt and can be visualized by plotting high frequency split node
questions. The base filters banks are however manually designed.
\textbf{Bottom left} is the block diagram for a typical convolutional neural
network which learns the filters as well as the discriminative features.
Visualizing them however needs additional trained filters or optimization
steps. \textbf{Bottom right} is our proposed TemplateNet which also performs
end-to-end learning. Plotting the output of a template layer is a simple way to
visualize the learnt features (figure~\ref{fig:templateResponse}).}}
    \label{fig:blockDiagram}
\end{figure}

Figure~\ref{fig:blockDiagram} shows the block diagram representations of
different
approaches. The top left pane shows the block diagram for
LineMod~\cite{hinterstoisser2011multimodal} which consists of four blocks. In
LineMod, given the input, it is first filtered using a manually designed
orientation filter bank. These filtered responses are then matched with object
template \textit{feature masks} which are manually designed and tuned to
highlight sparse \textit{discriminative} features (such as edges or corners) for
each template of an object. Finally, either using a learnt classifier or a
scoring function, the input is classified as foreground or background. As these
blocks are manually designed we have a good understanding of this system and
how it learns to recognize different objects. 

The top right pane shows the block diagram of the slRF-based instance
recognition system of~\cite{conf/eccv/BondeBC14} where the discriminative
features (or feature masks) are learnt across different templates. Visualizing
the features used by the split nodes gives an understanding of the learnt
feature masks. Bottom left pane shows the block diagram of a typical
feed-forward CNN.
Here an end-to-end system is used to learn the filter banks as well. Although
methods have been proposed to visualize these deep
networks~\cite{zeiler2010deconvolutional,simonyan2013deep} they either require
redundant layers or additional optimization steps.

\begin{figure*}[t]
    \centering   
\includegraphics[width=1\linewidth]{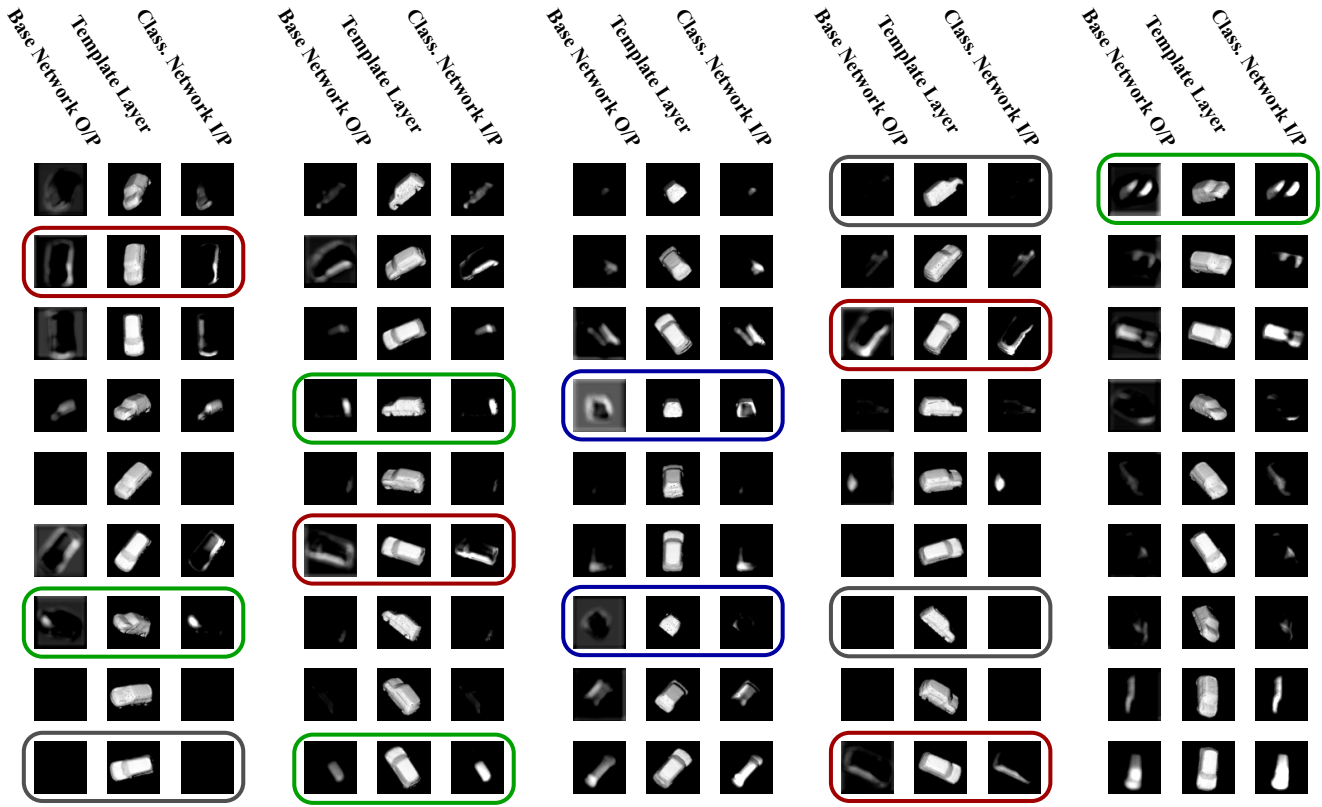}
    \caption{{Interpreting templateNet: Learnt feature mask responses from our
proposed templateNet for class Mini. We highlight some intermediate features
learnt by the first network. In red we highlight edge orientation features, in
green we highlight surface orientations. Lack of orientation, i.e high for
everything but a specific orientation is shown in blue. Gray shows voids, i.e no
features learnt. Here we show the responses only corresponding to one surface
normal orientation. Similar observations were made with other channels.}} 
    \label{fig:templateResponse}
\end{figure*}

In templateNet, shown in the bottom right pane of figure~\ref{fig:blockDiagram},
we split the deep neural network into two separate networks with an additional
template layer inserted between them (figure~\ref{fig:architecture}). The
base network learns the orientation filters and the feature masks for templates.
Ideally, we want these feature mask to be sparse and contain only the
discriminative features. However, rather than enforcing sparsity on the feature
masks we use the template layer as a \textit{sparsity inducing module}. The
weights of a template layer correspond to different template views of the
object. As these templates have structured shapes they force the template layer
output to also contain structure in their sparse
activation (figure~\ref{fig:architecture} and~\ref{fig:templateResponse}). This
is in contrast to~\cite{poultney2006efficient} which does not enforce any
spatial structure to their sparse feature maps. Finally, the classification
network uses these sparse maps as input to make the predictions. 
Visualizing the template layer output is an intuitive way to understand the
learnt features. Figure~\ref{fig:templateResponse} shows these learnt responses
for each template in the template layer of object class \textit{Mini} in the
dataset Desk3D~\cite{conf/eccv/BondeBC14}. The first column in each pane shows
the learnt feature masks which is the output of base network. The second column
shows the templates used in the fixed template layer. The final column is the
rectified linear output of the element wise product between first two columns
and is the input for the classification network. We also highlight some of the
intuitive features learnt by this network such as edge orientation (shown in
red) and surface orientation (shown in green). For clarity, in each row, the
input used in the base network was the same as that in the corresponding
template layer. Similar results were observed with different inputs. In the next
section we explain the various aspects of training the templateNet.

\subsection{Depth-based Instance recognition}
Given the depth image of a scene, various maps such as height from ground plane,
angle with gravity vector, curvature~\cite{gupta2014learning} can be computed to
be used as input. However, here we follow existing methods on instance
recognition which show state-of-the-art performance using surface normal
features~\cite{hinterstoisser2011multimodal,conf/eccv/BondeBC14}. These features
can be efficiently computed using depth maps.
We normalize each channel of the surface normal (x,y,z) in the range of $[0,1]$
and use it as our input. The template layers weights are also given by
the surface normals of the corresponding template view. We use 45 different
viewpoints (5 along yaw, 3 along pitch and 3 along roll) for each channel giving
a total of $135$ ($45\times3$) templates in the template layer.

\subsubsection{OrthoPatch}\label{sec:orthoPatch}
As we use calibrated depth sensors we can leverage the physical dimensions
encoded in a depth image to avoid searching over multiple window sizes. For this
reason, using the camera calibration matrix, we compute the world co-ordinates
from a depth image. We then perform orthogonal projections of the scene to get
its
\textit{orthoPatch}. The resulting orthoPatch encodes the physical dimensions to
a fixed scale thus removing the need to search over multiple scales.

\subsubsection{Training}\label{sec:train}
During training, we use a 3D model of the object and project it from different
viewpoints. We simulate background clutter by randomly adding other
objects along with floor before computing its orthoPatch. To improve robustness
we also add random shifts. We crop this to $128
\times 128$ dimension and simulate $27000$ such views to get the foreground (fg)
training data. For background (bg) training data we use $128 \times 128$
orthopatches from frames of video sequences containing random clutter to obtain
the final training set $\mathbb{X}$.


Most methods in depth-based instance recognition rely on
ICP~\cite{besl1992method}
or its robust versions to get the final pose estimate of an object. Their
primary focus is on having a good prediction for ICP initialization. We use a
similar scheme and treat pose as a classification task rather than regression.
To this end, we uniformly quantize object viewpoints into 16 pose classes with
the goal of predicting the closest pose class. We thus have a $16+1$ (poses +
background)
\textit{pose classification} task. As a test-object pose could be between two
\textit{quantized pose classes} rather than forcing the network to assign a
specific quantized pose we use \textit{soft-labels}~\cite{budvytis2010label}.
This helps to better explain simulated poses that are not close to any single
pose class. Soft labels for each simulated example are assigned based on the
deviation of its \textit{canonical rotational matrix} to the identity
matrix~\cite{journals/jmiv/Huynh09}. For example, if $R_i^j$ represents the
canonical rotation matrix which takes the $i^{th}$ simulated view to the
$j^{th}$ quantized pose, then its distance is given by: $ d_i^j = \|I -
R_i^j\|_F,$ where $I$ is a $3\times3$ identity matrix and $\|.\|_F$ is the
Frobenius norm. The soft labels are then given by: $ l_i^j = \exp(-{d_i^j}^2), j
\in \left\{1,...,16\right\}$. The final label vector $l_i$ is normalized to get
the soft labels.

During testing, the sum of predicted labels across all quantized pose classes is
used to estimate the foreground probability ($p_{fg}$). However, as we show
later in our experiments, using this as the labels for training is not an
optimal approach for detection. This is because a model learnt with only the
pose classification cost does not explicitly maximize foreground probability. 

In order to address both tasks of fg-bg and pose
classification we take inspiration from the work of transforming
auto-encoders~\cite{hinton2011transforming}. We use a two-headed model with one
head predicting the fg-bg probability which is invariant over the viewing
domain. The other head predicts pose class probability which varies uniformly
over the viewing domain and is similar to their \textit{instantiation
parameters}.
This explicitly introduces the fg-bg objective into the cost function. The final
cost function is given by the cross entropy as:
\begin{equation}\label{eq:mixCostFunc}
\begin{array}{l}
  \textbf{L}(\theta) = \frac{1}{N}\sum_{n=[1,\cdots,N]}-\textbf{L}^n  \\
  \textbf{L}^n = \sum_{i = [1,2]}
\textbf{y}_i^c log(\textbf{p}_i^c(\theta)) + \lambda \sum_{j = [1,17]}
\textbf{y}_j^p
log(\textbf{p}_j^p(\theta))
\end{array}
\end{equation}
Here the first term is the cross entropy for fg-bg classification with $y_i^c$
taking binary values (superscript $c$ indicating fg-bg) and the second term is
for pose classification with $y_j^p$ given by the soft labels (superscript $p$
indicating pose). $\theta$ are the parameters of the network. We suppress the
superscript $n$ from the second equation for clarity. $\lambda$ acts as the
reweighing term to normalize the two cost functions. The resulting model with a
mixed objective outperforms other single headed models that consider fg-bg and
pose separately.

\subsubsection{Learning}\label{sec:learn}
The individual components in templateNet can be formulated as:
\begin{itemize}
 \item \textit{Base network} can be simplified as a single
convolutional layer with non-linearity: $\hat{\textbf{z}} =
\phi(\widehat{W}^T\textbf{x} + \widehat{\textbf{b}})$
 \item \textit{Template layer} can be formulated as a multiplicative
layer with non-linearity: $ \textbf{z} = \phi(\hat{\textbf{z}}\cdot T)$
\item \textit{Classification network} can be approximated as a softmax
layer: $ \textbf{p} = \sigma(W^T\textbf{z} + \textbf{b})$
\end{itemize} 
where, $\textbf{x}$ is the input, $\widehat{W}$,$W$ and
$\widehat{\textbf{b}}$,$\textbf{b}$ are the filter weights and biases, $T$ are
the templates (or scaling factors), $\phi$ is the non-linearity (ReLU), $\sigma$
is the softmax function and $\textbf{p}$ is the final predicted probability.
Thus the parameters of the network are $\theta
=\{\widehat{W},W,\widehat{\textbf{b}},\textbf{b},T\}$. We use the mixed cross
entropy cost (equation (\ref{eq:mixCostFunc})) as our cost function for
training. We start from the classification network where the predicted
probability
is given as:
\begin{equation}
 \textbf{p}_i = \frac{e^{-[W_i^T\textbf{z}+\textbf{b}_i]}}{\sum_j
e^{-[W_j^T\textbf{z}+\textbf{b}_j]}} =
\frac{e^{-[W_i^T\textbf{z}+\textbf{b}_i]}}{\textbf{Z}}
\end{equation}
Subscripts $i$ and $j$ are used to index the $i^{th},\, j^{th}$ components of
a vector (bold lower case variables) or the $i^{th},\, j^{th}$ columns of a
matrix (upper case variables). Using the cross entropies per example
term ($\textbf{L}^n$) from (\ref{eq:mixCostFunc}) (summation over
$n$ is independent of $\theta$), the partial derivatives for
classification network with respect to fg-bg cost are given by:
\[
 \frac{\partial \textbf{L}^n}{\partial W_{i}^c} = \textbf{y}_{i}^c \left[
\textbf{z} -
\frac{\textbf{z}e^{-[W_{i}^{c\;T}\textbf{z}+b_{i}^c]}}{\textbf{Z}^c}
\right]
\]
and
\[
 \frac{\partial \textbf{L}^n}{\partial b_{i}^c} = \textbf{y}_{i}^c \left[ 1 -
\frac{e^{-[W_{i}^{c\;T}\textbf{z}+b_{i}^c]}}{\textbf{Z}^c}
\right]
\]
The derivatives with respect to pose cost are the same with the additional
reweighing term $\lambda$. Partial derivatives with respect to the input
$\textbf{z}$ is given by:
\[
 \frac{\partial\textbf{L}^n}{\partial \textbf{z}} = \sum_{i=[1,2]}
\textbf{y}_{i}^c \left[ W_{i}^c - \frac{\sum_{k=[1,2]}
W_{k}^ce^{-[W_{k}^{c\;T}\textbf{z}+b_{k}^c]}}{\textbf{Z}^c}\right] + 
\]
\[
 \lambda \sum_{j=[1,17]} \textbf{y}_{j}^p \left[ W_{j}^p
- \frac{\sum_{l=[1,17]}
W_{l}^pe^{-[W_{l}^{l\;T}\textbf{z}+b_{l}^p]}}{\textbf{Z}^p}\right]
\]
Using chain rule we compute the partials derivatives with the template
layer as:
\begin{equation}~\label{eq:tempUpdate}
  \frac{\partial\textbf{L}^n}{\partial T} = \frac{\partial\textbf{L}^n}{\partial
\textbf{z}} \frac{\partial\textbf{z}}{\partial T} =
\frac{\partial\textbf{L}^n}{\partial \textbf{z}} \;\phi ' (\hat{\textbf{z}}\cdot
T) \hat{\textbf{z}} \;\;\;\; 
\end{equation}
and
\[
\frac{\partial\textbf{L}^n}{\partial \hat{\textbf{z}}} =
\frac{\partial\textbf{L}^n}{\partial
\textbf{z}} \frac{\partial\textbf{z}}{\partial \hat{\textbf{z}}} =
\frac{\partial\textbf{L}^n}{\partial \textbf{z}} \;\phi ' (\hat{\textbf{z}}\cdot
T) T
\]
where, $\phi'$ is the derivative of ReLU which is equal to one for
positive values and zero otherwise. Finally the partial derivatives with respect
to masked response network parameters are given by:
\[
 \frac{\partial\mathbf{L}}{\partial \widehat{W}} =
\frac{\partial\mathbf{L}}{\partial\hat{\textbf{z}}}\frac{\partial\hat{\textbf{z}
}}{\partial \widehat{W}} =
\frac{\partial\textbf{L}^n}{\partial\hat{\textbf{z}}}\;
\phi'[\widehat{W}^T\textbf{x}+\widehat{\textbf{b}}]\;\textbf{x} 
\]
\[
\frac{\partial\mathbf{L}}{\partial \widehat{\textbf{b}}} =
\frac{\partial\mathbf{L}}{\partial\hat{\textbf{z}}}\frac{\partial\hat{\textbf{z}
}}{\partial \widehat{\textbf{b}}} =
\frac{\partial\textbf{L}^n}{\partial\hat{\textbf{z}}}\;
\phi'[\widehat{W}^T\textbf{x}+\widehat{\textbf{b}}]
\]

In the current work we do not update the $T$ and set
$\frac{\partial\textbf{L}^n}{\partial T} = 0$. However in future we could use
(\ref{eq:tempUpdate}) to update the template layer weights as well. To
enforce sparsity an additional term such as penalizing $||T_i||_1$ might be
needed. As the template layer weights are fixed it does not result in any
additional training parameters. We train the templateNet in an end-to-end way
similar to a typical CNN.
\begin{figure}[t]
    \centering   
\includegraphics[width=1\columnwidth]{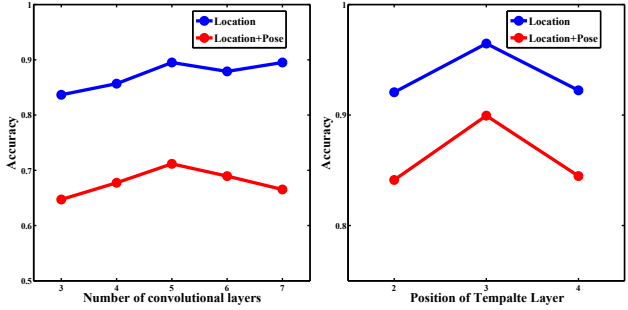}
    \caption{{Location and location+pose accuracies for: \textbf{left} pane
shows the performance with increasing convolutional layers for an
object from Desk3D and \textbf{right} pane shows the performance for different
positions of the template layer in templateNet.}}
    \label{fig:explore}
\end{figure} 

\subsection{Analysis}
\begin{figure}[t]
    \centering   
\includegraphics[width=1\linewidth]{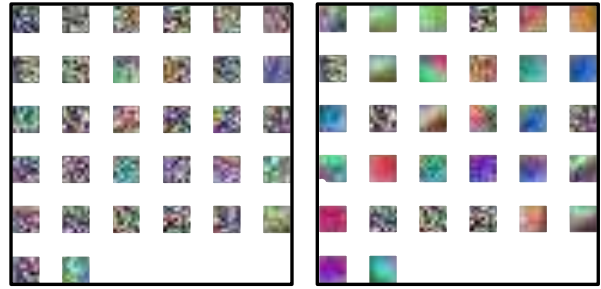}
    \caption{{First layer filters of different networks. \textbf{Left}
pane shows the filters learnt by a five-layered network using
surface normals from depth images which look random and lack structure.
\textbf{Right} pane compares the filters learnt by the templateNet. Due to
stronger regularisation in templateNet most of the resulting filters have better
structure.}}
    \label{fig:firstLayer}
\end{figure}

In this section we analyse the effect of different settings of templateNet and
use them as a guide for further experiments in section~\ref{sec:chpt3_exp}.\\
\noindent\textbf{Number of Convolutional layers:} We test the effect of
increasing the number of convolutional layers using the
two headed network for a typical CNN. Figure~\ref{fig:explore} (left pane)
compares the performance for class Kettle in Desk3D. We use the same training
protocol of hardmining for all settings, i.e starting with a random subset of
training data $\mathbb{X}_s \in \mathbb{X}$ followed by hard-mining after every
5-10 epochs for a total of 50 epochs. From the plot we observe the performance
to start overfitting after five layers. Following these results we use five
convolutional layers with two fully-connected layer as our base model for all
other experiments.\\
\noindent\textbf{Depth of Template layer:} As a template layer is essentially a
multiplicative layer it can be placed between any two convolutional layers. We
experiment by placing the template layer at different depths in the five-layered
CNN. Figure~\ref{fig:explore} (right pane)
compares the resulting performance. Having more convolutional layers before the
template layer leads to a
larger non-linearity and hence more complex features to be learnt resulting in
an improved performance. However as the template layer moves further from the
input its regularisation effect on the initial filters reduces and the
performance degrades. We found the templateNet performs best when the template
layer is placed after the third convolutional layer in the five layered CNN.
Figure~\ref{fig:architecture} (bottom pane) shows the final architecture of
templateNet.\\
\noindent\textbf{First layer filters:} The left pane of
figure~\ref{fig:firstLayer} shows the learnt filters in the first layer of our
five-layered CNN. The noisy and unstructured first
layer filters in the five-layered CNN can be accounted for by two factors: a)
the model is over-parametrized; b) the model is not well regularised. However,
from figure~\ref{fig:explore} (left pane) we observe that the test error
decreases with an increase in number of layers (or parameters). This suggests
that over-parametrization is not the primary cause and that the model is not
well regularised. In comparison, the sparsity induced by the template layer
regularises the templateNet and forces the filters to explain the data better
making them structured and less noisy~\cite{memisevic2013learning}. We
observe this effect in figure~\ref{fig:firstLayer} (right pane).

\section{Experiments and Results}\label{sec:chpt3_exp}
Several datasets exist in the literature for testing instance recognition
algorithms~\cite{conf/icra/LaiBRF11,conf/iccvw/BrowatzkiFGBW11,
conf/icra/TangMSA12}. Of these we choose the Desk3D~\cite{conf/eccv/BondeBC14}
and the ACCV3D~\cite{hinterstoisser2013model} datasets. Unlike other datasets
the Desk3D dataset contains separate scenarios to test the performance of the
recognition algorithms under different challenges of similar looking
distractors, pose change, clutter and occlusion. The controlled test cases
allows us to better analyse each algorithm to estimate their performance under
real world conditions. However ,this dataset is of a limited size and for this
reason we also experiment with the ACCV3D dataset which is the largest publicly
available labelled dataset covering large range of pose variations with
clutter and multiple object shapes.\\
\noindent\textbf{Benchmarks}: Two different benchmarks are used to quantify and
compare different settings. We use the state-of-the-art slRF
method~\cite{conf/eccv/BondeBC14} together with the depth-based
LineMod~\cite{hinterstoisser2011multimodal} as our base benchmarks. Since
LineMod learns templates for RGB and depth separately, removing one modality
does not affect the other. For reference we also report results from depth+HoG
based DPM~\cite{felzenszwalb2008discriminatively}.\\
\begin{figure*}[t]
  \begin{center}
  
\includegraphics[width=1\linewidth]{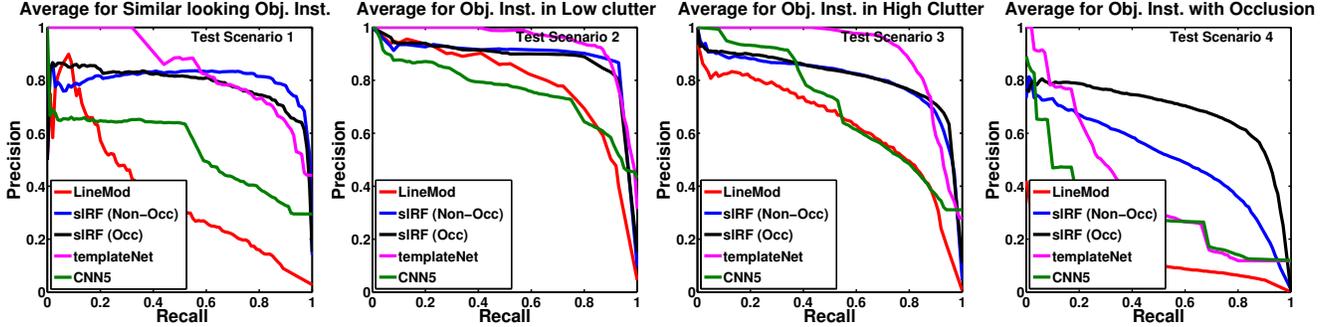}
  \end{center}  
     \caption{Precision-Recall curves for pose+location for different test
scenarios in
Desk3D. TemplateNet learns features directly from the input data and is more
confident in its predictions. The template layer introduces sparsity and
regularises the network. The resulting architecture improves performance over
traditional CNN and outperforms previous state-of-the-art methods of LineMod and
 in three out of four challenges ((a) similar objects, (b) large pose variations
and (c) clutter). As templateNet does not encode the occlusion information it
fails on occluded scenes (d).} 
  \label{fig:Desk3DScenarios}
\end{figure*}
\noindent\textbf{Testing Modality}: We follow the same testing modality
as~\cite{conf/eccv/BondeBC14}. We consider an object to be correctly
localized if the predicted centre is within:$\frac{max(w,d,h)}{3}$
radius of the ground truth. For pose classification we
consider pose to be correctly classified if the predicted pose class
$p_{ps}^{*}$ given by: $ p_{ps}^{*} = \operatorname*{arg\,max}_i p_{ps}^i $ i.e,
largest pose probability is either the closest or second closest quantized pose
to the ground truth.\\

\begin{table*}[t!]
\begin{center}
\begin{tabular}{|p{1.5cm}||p{1.5cm}|p{1.5cm}|p{1.5cm}|p{1.5cm}|p{1.5cm}|p{
1.5cm}|p{1.6cm}||p{1.6cm}|}
\hline

 &
DPM~\cite{felzenszwalb2008discriminatively} &
LineMod &
slRF~\cite{conf/eccv/BondeBC14}&
CNN3 &
CNN3 &
CNN5 &
templateNet &
templateNet\\

Object &
(D-HoG)&
\cite{hinterstoisser2011multimodal} &
(Normals)&
(Pose)& 
(Mix)&
(Mix)&
(Mix-45)&
(Mix-128)\\

\hline\hline
Face & 44.74 & 66.60 & 69.90 & 67.22 & 62.89 & 67.84 & \textbf{70.72} &
72.63
\\
Kettle & 53.34 & 60.49 & 83.75 & 72.31 & 70.55 & 72.49 & \textbf{84.13}
& 89.07\\ 
Ferrari & 32.41 & 47.90 & 50.42 & 59.66 & 69.75 & 68.07 & \textbf{81.51}
& 80.11\\  
Mini & 30.64 & 55.09 & 63.74 & 36.54 & 57.74 & 60.53 & \textbf{77.96} &
76.78\\ 
Phone & 64.32 & 79.03 & \textbf{91.85} & 80.94 & 69.67 & 70.54 & 83.36
& 90.81\\ 
Statue & 70.29 & \textbf{84.76} & 81.97 & 39.22 & 50.37 & 55.95 & 83.09
& 84.92\\ 
\hline\hline
Average & 49.29 & 65.64 & 73.64 & 59.31 & 63.49 & 65.90 & \textbf{80.13}
& 82.39\\
\hline
\end{tabular}
\end{center}
\caption{{Location+Pose accuracies on non-occluded scenes of Desk3D. We compare
the templateNet with different settings of the convolutional neural network
(CNN) together with previous state-of-the-art methods. Using the mixed cost
function (mix) performs better than the pose+background cost (pose). Increasing
the number of convolutional layers further improves the accuracy. However as CNN
is not well regularised it still lags behind the state-of-the-art methods of
LineMod~\cite{hinterstoisser2011multimodal} and slRF~\cite{conf/eccv/BondeBC14}
(normal based). TemplateNet uses the template layer to enforce sparsity 
resulting in top performance in 4 out of 6 instances and the best overall
performance. Increasing the number of templates from 45 to 128 results in
further improvement but at a cost of large processing time. The results with
larger templates (Mix-128) are only quoted for reference. We do not use them for
other experiments. To avoid clutter we report only the location+pose accuracies
and not the location only accuracies. Similar results were observed with
location accuracies as well.
}\label{table:nonOccScenes}}
\end{table*}
\subsection{Experiments on Desk3D}\label{sec:desk3d}
The Desk3D dataset contains a total of six objects with 400-500 test scenes
each. The test scenes are obtained by fusing few consecutive frames ($\approx
5$) using~\cite{Rusu_ICRA2011_PCL}. Figure~\ref{fig:Desk3DScenarios} shows PR
curves for
the four different scenarios in Desk3D. Figure~\ref{fig:Desk3DScenarios} (a)
compares the performance on scenario 1 which consists of similar looking
distractors. As the templateNet performs feature learning it outperforms other
methods which use manually designed features. In
figure~\ref{fig:Desk3DScenarios} (b) we compare the performance for the low
clutter and high pose variations stetting (scenario 2). TemplateNet is more
confident for large pose variations and we observe a high precision for over
50\% recall rates. A similar improvement is observed even with large cluttered
background of scenario 3 (figure~\ref{fig:Desk3DScenarios} (c)). Due to a better
separation between foreground-background the templateNet is more confident even
with large cluttered background giving high precision at large recalls.

In all three scenarios (\ref{fig:Desk3DScenarios} a,b and c) the five layered
CNN performs the worst compared to other benchmarks indicating a need for better
regularisation. The templateNet achieves this using the sparsity inducing
template layer resulting in the best performance.

Table~\ref{table:nonOccScenes} lists the accuracies of different settings for
non-occluded scenes. For fairness we do not add dropouts in any of the
architectures. The use of a mixed cost helps improve performance over a
pose classification cost. This is because a model trained with the pose only
cost does not explicitly maximize the foreground probability. With an increase
in the number of layers the performance improves and saturates around five
convolutional layers (figure~\ref{fig:explore}). Using better regularisation our
templateNet outperforms all others in four out of six objects and achieves the
best overall accuracy. We also experiment with the width of the template layer
i.e, number of templates used. As the number of templates increases we increase
the
representation power/dimensions of template layer while still regularising the
network due to sparse activations. This leads to an improved performance of the
network seen in the last column of table~\ref{table:nonOccScenes} albeit at a
higher computational cost. For this reason we only report these results for
reference.

The only scenario where templateNet suffers is under occlusion
(figure~\ref{fig:Desk3DScenarios} (d)). Using occlusion information the slRF
outperforms templateNet when objects are partially
occluded. In future using occlusion information could help the templateNet
handle partially occluded scenes.
\begin{figure}[t]
  \begin{center}
  
\includegraphics[width=1\linewidth]{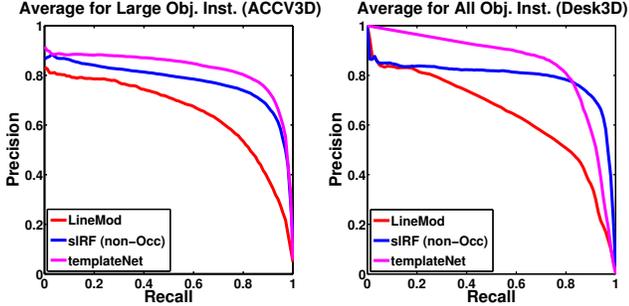}
  \end{center}  
     \caption{Precision-Recall curves for pose+location: \textbf{left} pane
shows the
performance on large objects (bounding box $> 1000 cm^3$) of ACCV3D dataset.
\textbf{Right} pane shows the average performance for non-occluded scenes of
Desk3D (Test scenarios 1,2 and 3).} 
  \label{fig:avgResults}
\end{figure}
\subsection{Experiments on ACCV3D}
The ACCV3D dataset contains $15$ different objects each with over $1100$ test
scenes covering a range of poses. However we only experiment with $9$ large
objects (bounding box $> 1000$cc). This is because due to low sensor resolution
the smaller $6$ objects do not have enough discriminative features and perform
poorly. This observations is also consistent with the results
of~\cite{conf/eccv/BondeBC14}. As our simulated training examples do not contain
any noise modelling our system is not robust to raw depth images which contain
boundary and quantization noise together with missing depth values. For this
reason we use the depth in-painting technique proposed
in~\cite{silberman2012indoor} to smooth the raw depth images and use them as our
input.
\begin{table}
\begin{center}
\begin{tabular}{|p{1cm}||p{0.75cm}|p{0.75cm}||p{0.75cm}|p{0.75cm}||p{0.75cm}|p{
0.75cm}|}
\hline
Object &\multicolumn{2}{c||}{LineMod~\cite{hinterstoisser2011multimodal}}
&\multicolumn{2}{c||}{slRF~\cite{conf/eccv/BondeBC14}} &
\multicolumn{2}{c|}{templateNet}\\
\cline{2-7}

Instance & L & L + P & L & L + P & L & L + P \\
\hline\hline

B.Vise & 83.79 & 75.23 & 87.98 & 86.50 & \textbf{96.79} &
\textit{\textbf{92.18}}\\

Camera & 77.19 & 68.94 & 58.20 & 53.37 & \textbf{83.76} &
\textit{\textbf{80.27}}\\

Can & 83.70 & 69.57 & 94.73 & 86.42 & \textbf{95.31} & \textit{\textbf{87.88}}\\

Driller & \textbf{94.70} & 81.82 & 91.16 & \textit{\textbf{87.63}} & 91.16 &
81.14 \\

Iron & 83.51 & 75.43 & 84.98 & 70.75 & \textbf{89.58} &
\textit{\textbf{82.64}}\\

Lamp & 92.91 & 80.93 & \textbf{99.59} & \textit{\textbf{98.04}} & 95.44 &
81.99\\

Phone & 77.72 & 70.47 & 88.09 & 87.69 & \textbf{94.05} &
\textit{\textbf{91.39}}\\

Bowl & 98.11 & 19.22 & \textbf{98.54} & 30.66 & 91.08 &
\textit{\textbf{32.15}}\\

Box & 63.37 & 27.69 & 95.21 & 63.53 & \textbf{95.29} & \textit{\textbf{72.47}}\\

\hline\hline
Average & 84.00 & 63.26 & 88.62 & 73.64 & \textbf{92.50} &
\textit{\textbf{78.01}}\\
\hline
\end{tabular}
\end{center}
\caption{{Location (L) and Location+Pose (L+P) accuracies on the ACCV3D dataset.
We report accuracies only for large objects (bounding box $> 1000$cc). Due to
the poor resolution of depth sensors small objects do not
perform well. We train a mixed cost templateNet with 45 templates. Our
end-to-end
learned network outperforms existing state-of-the-art methods in 6 out of 9
categories and gets the best overall performance.}\label{table:tum}}
\end{table}
Figure~\ref{fig:avgResults} (left pane) shows the average precision-recall
curves for
different methods and table~\ref{table:tum} reports the performance on the large
objects in ACCV3D dataset. Our end-to-end trained templateNet achieves the best
performance on 6 out of the 9 objects and the best overall performance. 

\section{Discussions}
In this section we discuss some of the limitations of the current architecture
and suggest possible directions to address these drawbacks.

In the current work, templateNet does not reuse the bottom level features
requiring one netwrok per object. This could be addressed by having
multiple template layers working in parallel with an additional group sparsity
penalty~\cite{huang2010benefit} on them. This would limit the number of template
layers being active while reusing the existing layers making it more efficient
and scalable.

The other drawback as shown in our experiments (section~\ref{sec:desk3d}) is its
poor performance in occluded scenes. With the use of spatial
transformers~\cite{jaderberg2015spatial} we could detect salient parts to
perform part based recognition to address this limitation. Nevertheless,
templateNet outperforms existing works on all other challenges of
instance recognition using only depth data.

\section{Conclusions}
We presented a new deep architecture called templateNet for depth-based object
instance recognition. The new architecture used prior knowledge of object shapes
to introduce sparsity on the feature maps. This was achieved without any
additional parametrization of training by using an intermediate template layer.
The sparse feature maps implicitly regularised the network resulting in
structured filter weights. By visualizing the output of a template layer we get
an intuition of the learnt discriminative features for an object. We derived the
weight updates needed to train the templateNet in an end-to-end manner. We
evaluated its performance on challenging scenarios of Desk3D as well as on the
largest publicly available ACCV3D dataset. We showed that the template layer
helped improve performance over a traditional convolutional neural network and
outperforms existing state-of-the-art methods. 
\section{Acknowledgements}
This research was supported by the Boeing Company. We gratefully acknowledge
Paul Davies for his inputs in making this work useful to the industry.

{\small
\bibliographystyle{ieee}
\bibliography{references}
}

\end{document}